\pdfoutput=1

\documentclass[11pt]{article}

\usepackage[preprint]{acl}

\usepackage{times}
\usepackage{latexsym}

\usepackage[T1]{fontenc}

\usepackage[utf8]{inputenc}

\usepackage{microtype}

\usepackage{inconsolata}

\usepackage{graphicx}

\usepackage{times}
\usepackage{latexsym}
\usepackage[T1]{fontenc}
\usepackage[utf8]{inputenc}
\usepackage{microtype}
\usepackage{inconsolata} 

\usepackage{amsmath}
\usepackage{amssymb}  
\usepackage{amsthm}
\usepackage{mathtools}

\usepackage{booktabs}
\usepackage{multirow}
\usepackage{multicol}
\usepackage{tabularx}
\usepackage{tabulary}
\usepackage{float}

\usepackage{graphicx}
\usepackage{bm}
\usepackage{xcolor}
\usepackage{xcolor,colortbl}

\usepackage{enumitem}
\usepackage{subfig}
\usepackage{listings}
\newcommand{\ours}{\textsc{OptAgent}\xspace}

\newcounter{note}

\usepackage[ruled,vlined,linesnumbered]{algorithm2e}
\SetKwComment{tcp}{\small\textcolor{gray}{$\triangleright$\,}}{}
\DontPrintSemicolon

\title{\ours: Optimizing Multi-Agent LLM Interactions Through Verbal Reinforcement Learning for Enhanced Reasoning}

\author{
Zhenyu Bi\textsuperscript{1}, Meng Lu\textsuperscript{1}, Yang Li\textsuperscript{2},
Swastik Roy\textsuperscript{3}, Weijie Guan\textsuperscript{1}, Morteza Ziyadi\textsuperscript{3}, Xuan Wang\textsuperscript{1}\\
\textsuperscript{1}Virginia Tech  \textsuperscript{2}College of William and Mary  \textsuperscript{3}Amazon Alexa AI \\
\{zhenyub, menglu, skjguan, xuanw\}@vt.edu, yli102@wm.edu, \{roswasti, mziyadi\}@amazon.com
}

\begin{document}
\maketitle
\begin{abstract}
Large Language Models (LLMs) have shown remarkable reasoning capabilities in mathematical and scientific tasks. To enhance complex reasoning, multi-agent systems have been proposed to harness the collective intelligence of LLM agents. However, existing collaboration structures are either predefined or rely on majority voting or round-table debates, which can suppress correct but less dominant agent contributions. Recent approaches model multi-agent systems as graph networks but optimize purely for agent performance, neglecting the quality of interactions. We hypothesize that effective agent communication is crucial for multi-agent reasoning and that debating quality plays a significant role. To address this, we propose $\ours$, a multi-agent verbal reinforcement learning algorithm that dynamically constructs and refines multi-agent collaboration structures. Our method defines action spaces and a feedback mechanism that evaluates communication robustness and coherence throughout the debate. The final decision is achieved through a majority vote over all the agents. We assess $\ours$ on various reasoning tasks, including mathematical reasoning, creative writing, scientific reasoning, and numerical sorting. Results demonstrate that our approach significantly outperforms single-agent prompting methods and state-of-the-art multi-agent frameworks on diverse tasks.

\end{abstract}

\section{Introduction}

Large Language Models (LLMs) have exhibited significant potential in reasoning across various downstream tasks, including elementary mathematical reasoning, and fundamental science reasoning \cite{GPT35, lm3, CotWei,sccot}. Despite these initial successes, existing methodologies necessitate meticulously crafted prompt strategies that are often fixed for certain tasks \cite{Yao2023TreeOT,Got24}. This approach lacks flexibility, as the users have to define different prompts under different scenarios, especially for complex reasoning tasks. A promising solution that mitigates the challenge is to explore multi-agent frameworks that capitalize on the strengths of LLM-based agents.
Researchers proposed many multi-agent reasoning frameworks that enable collaborative debates among multiple LLM agents \cite{Chan2023ChatEvalTB, liang2024encouragingdivergentthinkinglarge, Chen2023ReConcileRC, Wang2023ASO, Chen2023AutoAgentsAF}, which are akin to human group problem-solving scenarios. 

Despite these initial successes, existing multi-agent LLM reasoning methods often follow pre-defined or simple group chatting collaboration structures. For example, AutoGen \cite{Wu2023AutoGenEN} and ChatEval \cite{Chan2023ChatEvalTB} employs pre-defined collaboration structures; ReConcile \cite{Chen2023ReConcileRC} employs group discussion with confidence-based consensus decision; MAD \cite{liang2024encouragingdivergentthinkinglarge} employs group debate with a meta-summarizer as the decision-maker. These methods do not account for the varying interactions of differently profiled agents, nor do they optimize the sequence of communications to ensure the most effective information flow for specific tasks. As a result, correct but less dominant agent contributions could be overlooked. We believe the interaction schemas should be more flexible and further optimized for task-specific communication efficacy.

Recent trends in multi-agent collaboration emphasize using graph optimization techniques to enable flexible, task-adaptable coordination among agents, enhancing efficacy and scalability in complex environments. Specifically, GPT-Swarm \cite{Zhuge2024LanguageAA} conceptualizes the multi-agent framework as a computational graph. The inspiration is drawn from a "Society-of-Mind" concept and highlights the communication and collaboration among agents. For optimization, the authors use reinforcement learning to optimize the agent interactions. While previous methods show reasonable performance, they tend to overlook the agents' debate quality, an important aspect of a multi-agent framework. We hypothesize that the interaction quality between the agents should also play an important role in the optimization process. More specifically, we believe the optimization algorithms should also consider metrics like wording clarity and logical coherency apart from agent performance metrics. 


To tackle the above challenges, we propose $\ours$, an LLM-based Verbal Reinforcement Learning framework for Graph Optimization on multi-agent collaboration. The goal of $\ours$ is to find the most effective interaction patterns in a multi-agent collaboration graph. $\ours$ explicitly considers communication quality when identifying the most effective connections between agents. To refine the multi-agent collaboration structure, $\ours$ contains a feedback agent that evaluates the quality of the agent interactions and an action agent that updates the multi-agent collaboration graph based on the feedback. The final decision is achieved through a majority vote over all the agents. We evaluate $\ours$ on various downstream reasoning tasks, including mathematical reasoning, scientific reasoning, creative writing, and sorting tasks. Our experimental results demonstrate that $\ours$ significantly outperforms single-agent prompting methods and state-of-the-art multi-agent debating schemas on diverse reasoning tasks across various LLM families. We also present a case study to illustrate the efficacy of our framework.



\section{Related Work}

\paragraph{LLM Reasoning Prompting}
The field of large language models (LLMs) has seen significant advancements in recent years, particularly in the area of reasoning prompting. Various prompt engineering methods have been developed, aiming to improve large language models’ reasoning ability across various tasks and domains. Chain-of-thought (CoT) prompting \cite{CotWei} prompts the large language models (LLMs) to divide their reasoning process into smaller steps when solving a question, forming a chain of thoughts. Chain-of-thought self-consistency prompting \cite{sccot} improves on the CoT method by proposing different reasoning chains and ensembles on the final result. Tree-of-thought (ToT) prompting method \cite{Yao2023TreeOT} actively maintains a tree of thoughts, where each thought is a coherent language sequence that serves as an intermediate step toward problem-solving. Graph-of-thought \cite{Got24} further improves ToT by constructing a Directed Graph instead of a tree. LLMs can loop over a thought to refine it and aggregate thoughts or chains. There are also other X-of-thought prompting methods developed for various different downstream tasks and datasets \cite{pot,aot,myself,gcot}. Another notable contribution to the field is the systematic survey on prompting techniques by the Prompt Engineering Guide \cite{Schulhoff2024ThePR}. This survey categorizes various prompting methods and their applications, emphasizing the importance of prompt design in enhancing LLM reasoning. 

\paragraph{Multi-Agent Reasoning}
Recent advancements in large language model (LLM) multi-agent frameworks have garnered significant attention in the field of artificial intelligence. Studies such as \citet{Wu2023AutoGenEN,Chen2023AutoAgentsAF, menglu} have highlighted the impressive reasoning capabilities of LLMs, which have been leveraged to create autonomous agent systems that are capable of complex problem-solving and perform better than single agents. 

The question is how researchers can design effective multi-agent reasoning frameworks. There have been several studies and analyses on the efficiency and effectiveness of multi-agent debating systems over reasoning tasks \cite{Wang2023ASO,Wang2024RethinkingTB,Pezeshkpour2024ReasoningCI}. However, most of the interaction schemas and decision strategies are either pre-defined \cite{Wu2023AutoGenEN, Chan2023ChatEvalTB}, or follow a simple structure such as group debate, majority voting, summarizer decision, or a combination of the above strategies \cite{Chen2023ReConcileRC, liang2024encouragingdivergentthinkinglarge, Chan2023ChatEvalTB}. Recently, several researchers from KAUST proposed GPTSwarm \cite{Zhuge2024LanguageAA}, in which they suggest that the multi-agent system can be considered as a graph network and thus their interaction patterns can be optimized by optimization algorithms. They also conduct individual optimizations on agents by conducting prompt optimization. However, their optimization is heavily performance-oriented, overlooking the debating quality of the agents. This is something that should also be considered in LLM free generation.

\section{$\ours${} Framework}
\begin{figure*}[t]
\includegraphics[width=15.5cm, height=6cm]{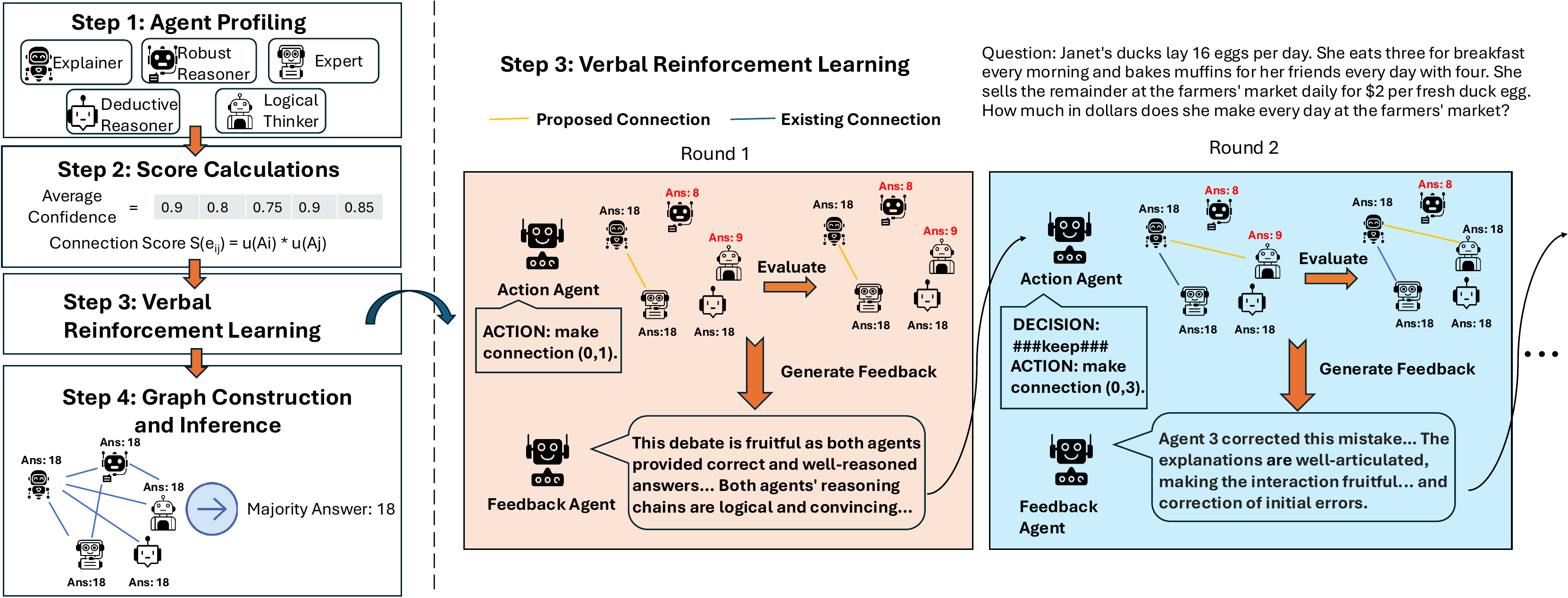}
\centering
\caption {Overview of $\ours$ framework. The overall pipeline is on the left side; an example process for verbal reinforcement learning is shown on the right.}
\label {Fig2}
\end{figure*}

\subsection{Problem Definition}
Given a problem $P$, and $N$ LLM agents $A_1, A_2, ..., A_N$, our goal is to find the answer to question $P$. We achieve this goal through using LLMs as agents to conduct logical reasoning and structured discussions. Each agent is a distinctly prompted LLM capable of generating the answer and the corresponding CoT reasoning process. 

\subsection{Framework Overview}
In our setting, we view the multi-agent collaboration framework as a graph. Each agent is a node in the graph, denoted by $A_i$; the communications between agents are the edges, denoted by $e_{ij}$. We hypothesize that the interaction quality will be different for differently profiled agents, and the best connection order would allow the best information propagation pattern for a particular task. The goal of $\ours$ is to optimize the connections between the agents and improve the overall performance of the multi-agent collaboration framework. 

In our verbal reinforcement learning process, we design two meta agents, $LLM_{reflect}$ and $LLM_{act}$. which handle reflection and action processes, respectively. The training process involves selecting connections based on probability scores and updating them through reinforcement learning. Finally, a majority voting strategy is used to determine the final answer after executing the graph.

\subsection{Initial Graph Setup}
\label{sec33}
\paragraph{Agent Profiling and Force Decoding}
Given a group of LLM agents $A_1,...A_i$, we ensure similar but different reasoning by assigning the agents with the same baseline reasoning prompt but different agent profiles in system prompts (see Appendix \ref{appendix:prompt}). The seven agent profiles were manually crafted to reflect common reasoning strategies found in human problem-solving, such as deductive logic, intuition, and domain expertise. 
For the 3-agent and 5-agent scenarios, we randomly select 3 and 5 profiles from the proposed profiles, respectively. To promote versatility, we force the model to generate three different outputs for each agent profile and randomly choose one of the outputs as its initial answer to the input question.

\paragraph{Connection Initialization}
Given a group of agents $A_1,...A_i$, and possible connections between the agents $e_{12},...,e_{ij}$, we first get the group of utility scores $u(A_i)$, which is the average self-evaluated confidence score given by the agent $A_i$ for the given task. We first randomly sample ten problems from the dataset, collect the confidence score from each agent on each question, and then calculate the average confidence score $u(A_i)$.

Then, we calculate the connection score of an edge, $s(e_{ij})=u(A_i)*u(A_j)$, which is determined by the utility score of the two connecting nodes. We will update the connection scores during the reinforcement learning process. Based on all of the connection scores, we assign the probability, $p(e_{ij})=\frac{s(e_{ij})}{\sum s(e_{ij})}$ to each connection $e_{ij}$, which is the proportion of the connection score $s(e_{ij})$ to the sum of the connection scores. The probabilities will serve as selection references in the first epoch of our training process.

\subsection{Verbal Reinforcement Learning}
Inspired by the Reflexion framework \cite{Shinn2023ReflexionLA}, we design an LLM self-controlled verbal optimization for graph generation. First, we design two meta agents: $LLM_{reflect}$ and $LLM_{act}$. We also create a set of action spaces that $LLM_{act}$ can choose from to alter the current graph network.

\paragraph{Reflection} \textbf{$LLM_{reflect}$} is responsible for generating reflection text after $LLM_{act}$ makes a connection between two agents $(A_i, A_j)$. Here, a 'connection' means initiating direct communication between two agents, prompting them to exchange their initial reasoning and answers, debate their points of view, and revise their reasoning based on the exchange. To generate the feedback, $LLM_{reflect}$ takes in the reasoning arguments of $A_i$ and $A_j$ before and after the interaction process. Then, the reflection text is passed on to $LLM_{act}$ to guide its decision-making process. Specifically, the feedback that $LLM_{reflect}$ generates is determined by two criteria:
\begin{itemize}[leftmargin=*]
    \item \textbf{Criterion 1}: Both agents should answer the question correctly after making the connection; 
    \item \textbf{Criterion 2}: Agents should be logical and coherent in their reasoning process. 
\end{itemize}
For the first criterion, $LLM_{reflect}$ checks whether the connection helps agent $A_i$ and $A_j$ with answering the question. If both agents got the answer correct, then $LLM_{reflect}$ will give positive feedback. For the second criterion, $LLM_{reflect}$ checks whether the logical chains are sound and valid. If both agents demonstrate good reasoning quality during the interaction process after seeing each other's reasoning, $LLM_{reflect}$ will give out good feedback. Otherwise, $LLM_{reflect}$ will have negative feedback on the connection $(A_i, A_j)$. Detailed instruction prompts for $LLM_{reflect}$ are provided in Appendix \ref{appendix:prompt}.

\paragraph{Action} \textbf{$LLM_{act}$} is responsible for conducting actions at each step, from the pre-defined action pool:
\begin{itemize}[leftmargin=*]
    \item Make a connection between the two agents $(A_i, A_j)$ to initiate debate;
    \item Keep a previously made connection $(A_n, A_m)$;
    \item Delete a previously made connection between the two agents $(A_n, A_m)$ to prohibit debate.
\end{itemize}
After $LLM_{act}$ receives the verbal feedback, it will make a decision to keep or delete the previously made connection. For instance, if $LLM_{act}$ decided to make a connection $(A_i, A_j)$ but consequently received negative feedback in this round, then $LLM_{act}$ would remove the connection. We decrease their connection score $s(e_{ij})$ for removed connections. If $LLM_{act}$ receives positive feedback, it will keep the connection $(A_i, A_j)$ in the graph, and we increase the connection score $s(e_{ij})$. Before deciding whether or not to keep the current edge, $LLM_{act}$ would also look back at the feedback history of the current edge in previous rounds. 

After the decision, $LLM_{act}$ makes a connection that hasn't been explored during the current training epoch. The result of the newly created connection will be evaluated and passed on to $LLM_{reflect}$ for the next round of reflection text generation.

\subsection{Training Process}
To start the Reinforcement learning process, we perform weighted random sampling to select a connection $(A_i, A_j)$ based on the probability score of the connections. At later epochs, $LLM_{act}$ is responsible for choosing a connection $(A_i, A_j)$. After $LLM_{act}$ takes action, we execute the debate process between $A_i$ and $A_j$, and then pass the results to $LLM_{reflect}$ for feedback, which is then given to $LLM_{act}$ for decision-making. We update the connection score $e_{ij}$ after $LLM_{act}$ has decided whether to keep the connection $(A_i, A_j)$. The connection score $s(e_{ij})$ is increased by $\alpha* \hat{s}(e_{ij})$ if $LLM_{act}$ chooses to keep it and decreases otherwise, where $\alpha$ is the learning rate we set, and $\hat{s}(e_{ij})$ is the current connection score of connection $e_{ij}$. We repeat the above process in the current epoch until every connection is visited once for an update. The pseudocode algorithm is provided in Algorithm \ref{alg_train} in the Appendix.

\subsection{Inference Process}
After the framework is trained with the connection weights updated, we construct the final graph before doing inference. Connections with higher scores are established first. The construction process continues until all agents have been visited. We consider the information flow within the graph as complete when each agent $A_i$ has interacted with at least one other agent $A_j$. The final decision is determined using a majority voting strategy as the final answer $Ans_{final}=mode(Ans_1, ..., Ans_n)$, where $Ans_1, ..., Ans_n$ are answers provided by different agents in the graph. The pseudocode algorithm is provided in Algorithm \ref{alg_inference} in the Appendix.

\section{Experiments}
\begin{table*}[t]
\centering
\scriptsize
\begin{tabulary}{\textwidth}{p{0.8cm}CCCCCCCC}
\toprule
\textbf{Model} & \textbf{Prompt Class} & \textbf{Framework Type} & \textbf{GSM8K} & \textbf{AdvGSM-M3} & \textbf{AdvGSM-M2} & \textbf{AdvGSM-M1} & \textbf{GSM-PLUS} & \textbf{MATH} \\
\midrule
\multirow{19}{*}{\parbox{1.5cm}{\textbf{GPT-3.5-turbo}}} & \multirow{3}{*}{\textbf{Single Agent}} & DirectIO  & 35.0 & 52.0 & 28.0 & 15.0 & 27.0 & 8.0 \\
 & & 0-Shot CoT & 73.0 & 87.0 & 75.0 & 30.0 & 59.0 & 22.0 \\
 & & ToT & 80.0 & 89.0 & 76.0 & 30.0 & 61.0 & 25.0 \\
 \cmidrule(lr){2-9}
 & \multirow{8}{*}{\textbf{3-Agent}} & Simple Debate & 77.0 & 90.0 & 79.0 & 31.0 & 62.0 & 25.0 \\
 & & GPT-Swarm & 79.6 & \textbf{91.3} & 80.6 & 33.6 & 63.0 & 28.0 \\
 & & ReConcile & 80.6 & 90.3 & 80.0 & \textbf{34.3} & 63.6 & \underline{29.0} \\
 \cmidrule(lr){3-9}
 & & \textbf{$\ours$} & \textbf{81.3} & \underline{91.0} & \underline{81.3} & \underline{34.0} & \textbf{64.3} & \textbf{29.3} \\
 & & Without Interaction Quality & \underline{81.0} & 90.0 & 81.0 & 33.0 & \underline{64.0} & \underline{29.0} \\
 & & No Forced Sampling & 79.0 & 89.0 & 81.0 & 32.0 & 63.0 & \underline{29.0} \\
 & & Reconsider Minority & 78.0 & 88.0 & 81.0 & 30.0 & 61.0 & 28.0 \\
 & & Split Action Agents & \underline{81.0} & 90.0 & \textbf{82.0} & \underline{34.0} & \underline{64.0} & \underline{29.0} \\
 \cmidrule(lr){2-9}
 & \multirow{8}{*}{\textbf{5-Agent}} & Simple Debate & 78.0 & 91.0 & 82.0 & 33.0 & 62.0 & 30.0 \\
 & & GPT-Swarm & 81.3 & 92.6 & 85.3 & 35.3 & \underline{66.6} & 32.6 \\
 & & ReConcile & 82.3 & 93.6 & \textbf{86.3} & 36.3 & 66.3 & 33.3 \\
 \cmidrule(lr){3-9}
 & & \textbf{$\ours$} & \textbf{87.3} & \textbf{95.6} & 85.3 & \textbf{38.6} & 66.0 & \underline{34.6} \\
 & & Without Interaction Quality & 84.0 & 94.0 & 84.0 & 36.0 & 64.0  & 33.0 \\
 & & No Forced Sampling & 84.0 & 94.0 & 84.0 & 37.0 & 65.0 & 32.0 \\
 & & Reconsider Minority & 85.0 & \underline{95.0} & \underline{86.0} & 36.0 & \textbf{69.0} & \textbf{37.0} \\
 & & Split Action Agents & \underline{86.0} & \underline{95.0} & 85.0 & \underline{38.0} & 66.0 & 34.0 \\
\midrule
\multirow{4}{*}{\textbf{GPT-4o}} & \multirow{4}{*}{\textbf{5-Agent}} & Simple Debate & \underline{97.0} & \underline{98.0} & 85.0 & 42.0 & 86.0 & 41.0 \\
& & GPT-Swarm & \underline{97.0} & \underline{98.0} & \underline{87.0} & \underline{44.0} & \underline{88.0} & \underline{42.0} \\
& & ReConcile & \textbf{98.0} & \textbf{99.0} & \underline{87.0} & \underline{44.0} & \textbf{89.0} & \underline{42.0} \\
\cmidrule(lr){3-9}
& & \textbf{$\ours$} & \textbf{98.0} & 98.0 & \textbf{88.0} & \textbf{45.0} & \underline{88.0} & \textbf{45.0} \\
\bottomrule
\end{tabulary}
\caption{Main results table on Math Reasoning Task. The best-performing methods on each dataset under each number-of-agent scenario are bolded, and the second-best are underlined. The results below \ours represent the variants of \ours framework. The detailed setting and discussion are presented in Section \ref{ablsec}.}
\label{tab1}
\end{table*}

\begin{table*}[t]
\centering
\scriptsize
\begin{tabulary}{\textwidth}{CCCCCCC}
\toprule
\textbf{Multi-Agent Framework} & \textbf{GSM8K} & \textbf{GSM8K-M3} & \textbf{GSM8K-M2} & \textbf{GSM8K-M1} & \textbf{GSM-PLUS} & \textbf{MATH} \\
\midrule
 \textbf{3 GPT-3.5-turbo} & 82.0 & \textbf{91.0} & 82.0 & 34.0 & \textbf{65.0} & 29.0 \\
\midrule
 \textbf{1 LLaMa3.1 70B + 2 GPT-3.5-turbo} & 83.0 & 87.0 & \textbf{84.0} & \textbf{35.0} & 63.0 & 33.0 \\
\midrule
 \textbf{2 LLaMa3.1 70B + 1 GPT-3.5-turbo} & 84.0 & 83.0 & 73.0 & 34.0 & 61.0 & \textbf{34.0} \\
\midrule
 \textbf{3 LLaMa3.1 70B} & \textbf{92.0} & 71.0 & 56.0 & 26.0 & 62.0 & 33.0 \\
\bottomrule
\end{tabulary}
\caption{Mixture of Model Ablation Task. All the multi-agent frameworks are optimized with $\ours$.}
\label{tab_mix}
\end{table*}

\subsection{Experimental Setup}
\paragraph{Dataset and Tasks} We experiment $\ours$ on four downstream tasks: math reasoning, creative writing, science reasoning, and sorting. All experiments were tested on publicly available datasets. For the math reasoning task, we use two datasets: GSM8K \cite{Cobbe2021TrainingVT}, which contains grade school arithmetic questions, and MATH \cite{Hendrycks2021MeasuringMP}, which contains high school-level mathematical questions spanning six different fields. We also include two adversarial reasoning datasets that are built on GSM8K: AdversarialGSM \cite{Xie2024AdversarialMW} in which we will refer to as AdvGSM in Table \ref{tab_gpt}, and GSM-PLUS \cite{Li2024GSMPlusAC}. AdvGSM contains questions that are changed only in number magnitude, and have three levels of difficulties, with M3 being the easiest using same magnitude with GSM8K, and M1 being the hardest. For each of the reasoning datasets except AdvGSM, we randomly select 100 questions from the dataset for evaluation. For AdvGSM, we randomly select 100 questions from each magnitude for evaluation. For creative writing, we follow the setup in \cite{Yao2023TreeOT}, where we test on 100 examples. For sorting, we randomly generate 100 numerical sequences at length 8, 16, 32. 

\paragraph{Model and Implementation} We experiment the baselines and $\ours$ utilizing GPT-3.5-turbo \cite{GPT35}, GPT-4o \cite{GPT4}, and the LLaMa 3.1-70B model \cite{lm3}. We directly call model APIs for prompting. For all models, we set the temperature to 0.5, and $top_k$ to 1.0. All base agents are prompted with the 0-shot CoT prompt. For each dataset, we train $\ours$ on three randomly sampled data points and report the performance on randomly sampled evaluation sets. We run \ours three times and report the mean performance. We use majority voting as our final decision strategy and random choice when there is a tie. We provide a cost analysis under the 5-agent scenario in Appendix \ref{appendix:cost}.

\paragraph{Baselines} We compared \ours with six single-agent prompting methods and state-of-the-art multi-agent baseline methods as below:
\begin{itemize}[leftmargin=*]
    \item \textbf{Single Model Prompts} in which we include 3 prompts: \textbf{DirectIO}, where we ask the model for a direct answer without explanations; \textbf{0-Shot CoT}, where we ask the model to provide step-by-step reasoning without providing any demonstrating examples; \textbf{ToT}, where we follow \citet{Yao2023TreeOT} and implement their framework.
    \item \textbf{Simple Debate}, where we initiate several instances of non-profiled agents with the same 0-shot CoT prompt. The agents are provided with each other's reasonings and answers, and are asked to reflect on their own reasoning. We let models debate for 2 rounds and utilize a majority voting to decide the final answer.
    \item \textbf{GPTSwarm} \cite{Zhuge2024LanguageAA}, where we follow the original implementation. We train the framework using three randomly sampled data points from the dataset and report the performance. We run GPTSwarm three times and report the mean performance.
    \item \textbf{ReConcile} \cite{Chen2023ReConcileRC}, where we follow the original implementation, using GPT-3.5-turbo and GPT-4o models as backbone, respectively. We report their performance in mathematical reasoning datasets. We run ReConcile three times and report the mean performance.
\end{itemize}

\subsection{Evaluation Metrics}
\paragraph{Math and Science reasoning} We report the performance in terms of accuracy following prior benchmarks and papers. The datasets include GSM8K, AdvGSM, GSM-PLUS, MATH, ARC and GPQA. We report the detailed post-processing and evaluation description in the Appendix.

\paragraph{Creative Writing} We follow the metrics in \citet{Yao2023TreeOT} and report the performance in terms of Coherence score, which another GPT-4 model evaluates. We provide the evaluation prompt in Appendix \ref{appendix:prompt}.

\paragraph{Sorting} We follow the metrics in \citet{Got24} and report the performance in terms of error scope, defined by the sum of the number of wrongly sorted elements and missing elements.

\subsection{Main Results}

\paragraph{Math Reasoning} 
We compare $\ours$ with multi-agent simple debating baselines on Math Reasoning datasets in Table \ref{tab1}. The backbone LLMs  (i.e., the primary large language model underlying all agents) include GPT-3.5-turbo and LLaMa 3.1-70B.  $\ours$ performs better on the original datasets like GSM8K and MATH than the simple debating baselines, and significantly outperforms the single-agent baselines. The performance increase is more prominent in 5-agent scenarios compared with 3-agent scenarios. We also present the results of two adversarial datasets in column 5 to 8. $\ours$ demonstrates robustness in the adversarial math reasoning datasets, outperforming the baseline scheme and frameworks by a similar margin compared with the original datasets. 

We also conduct experiments on the mathematical datasets with GPT-4o as the backbone model. With enhanced reasoning ability, even the simple debating method performs near-perfectly on basic math reasoning datasets. We still see a slight performance increase using the multi-agent debating frameworks on more challenging datasets. 

\paragraph{Creative Writing} 
\begin{figure}[t]
    \centering
    \resizebox{.5\textwidth}{!}{
    \includegraphics{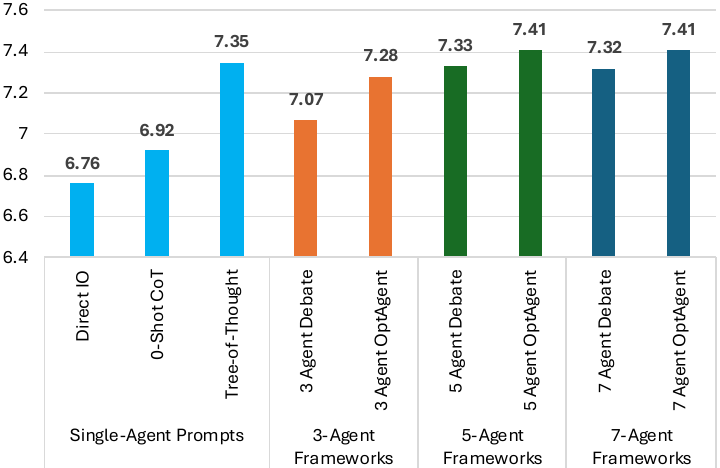}
    }
    \caption{Results on Creative Writing, measured in terms of coherence scores.}
    \label{fig3}%
\end{figure}
Results for creative writing task is reported in Figure \ref{fig3}. $\ours$ increase the coherence score by an average of at least 0.05 points across different settings under this task. Compared with Tree-of-Thought, which used a single model to explore different branches, \ours achieves slightly better performance. Increasing the number of agents only brings marginal performance improvement, and adding more agents from 5 to 7 does not seem to help with the performance of the multi-agent framework. 

\subsection{Ablation Study} 
\label{ablsec}
\begin{table*}[t]
\centering
\scriptsize
\begin{tabulary}{\textwidth}{CCCCCCCC}
\toprule
\textbf{Number of Agents} & \textbf{Framework Type} & \textbf{GSM8K} & \textbf{GSM8K-M3} & \textbf{GSM8K-M2} & \textbf{GSM8K-M1} & \textbf{GSM-PLUS} & \textbf{MATH} \\
\midrule
\multirow{4}{*}{\textbf{5-Agent}} & Random Initialization & 85.0 & 95.0 & 85.0 & 36.0 & 66.0 & 34.0 \\
\cmidrule(lr){2-8}
 & Uniform Initialization & \textbf{87.0} & 95.0 & \textbf{86.0} & 37.0 & \textbf{68.0} & \textbf{35.0} \\
\cmidrule(lr){2-8}
 & Confidence Scores & \textbf{87.0} & \textbf{96.0} & \textbf{86.0} & \textbf{38.0} & 67.0 & 34.0 \\
\bottomrule 
\end{tabulary}
\caption{Performance of \ours under different initialization methods for the connection scores.}
\label{tab_init}
\end{table*}

\begin{table*}[t]
\centering
\scriptsize
\begin{tabulary}{\textwidth}{CCCCCCCC}
\toprule
\textbf{Number of Agents} & \textbf{Framework Type} & \textbf{GSM8K} & \textbf{GSM8K-M3} & \textbf{GSM8K-M2} & \textbf{GSM8K-M1} & \textbf{GSM-PLUS} & \textbf{MATH} \\
\midrule
\multirow{3}{*}{\textbf{3-Agent}} & Simple Debate & 77.0 & 90.0 & 79.0 & 31.0 & 62.0 & 25.0 \\
& +Profiling & 82.0 (+5.0) & 90.0 (+0.0) & 82.0 (+3.0) & 33.0 (+2.0) & 64.0 (+2.0) & 29.0 (+4.0) \\
 & $\ours$ & 82.0 (+5.0) & 91.0 (+1.0) & 82.0 (+3.0) & 34.0 (+3.0) & 65.0 (+3.0) & 29.0 (+4.0) \\
 
\cmidrule(lr){2-8}
\multirow{4}{*}{\textbf{5-Agent}} & Simple Debate & 78.0 & 91.0 & 82.0 & 33.0 & 62.0 & 30.0 \\
 & +Profiling & 83.0 (+5.0) & 94.0 (+3.0) & 84.0 (+2.0) & 35.0 (+2.0) & 66.0 (+4.0) & 31.0 (+1.0) \\
 & $\ours$ & \textbf{87.0 (+9.0)} & 96.0 (+5.0) & \textbf{86.0 (+4.0)} & \textbf{38.0 (+5.0)} & 67.0 (+5.0) & \textbf{34.0 (+4.0)} \\
\cmidrule(lr){2-8}
 \multirow{3}{*}{\textbf{7-Agent}} & Simple Debate & 78.0 & 92.0 & 81.0 & 34.0 & 62.0 & 30.0 \\
 & +Profiling & 83.0 (+5.0) & 95.0 (+3.0) & 85.0 (+4.0) & 35.0 (+1.0) & 65.0 (+3.0) & 31.0 (+1.0) \\
 & $\ours$ & 85.0 (+7.0) & \textbf{98.0 (+6.0)} & \textbf{86.0 (+5.0)} & 37.0 (+4.0) & \textbf{68.0 (+6.0)} & 33.0 (+2.0) \\
\bottomrule 
\end{tabulary}
\caption{Performance of $\ours$ on GPT-3.5-turbo under 3, 5, and 7-agent scenarios. "Simple Debate" refers to agents debating without profiles and forced generation. "+Profiling" refers to debating with added profiles. $\ours$ contains both Profiling and Verbal Reinforcement Learning. We bold the best performing variant. The deltas stand for differences between variant from simple debate baseline.}
\label{tab_gpt}
\end{table*}

\paragraph{Training Convergence}
\begin{figure}[t]
    \centering
    \resizebox{.5\textwidth}{!}{
    \includegraphics{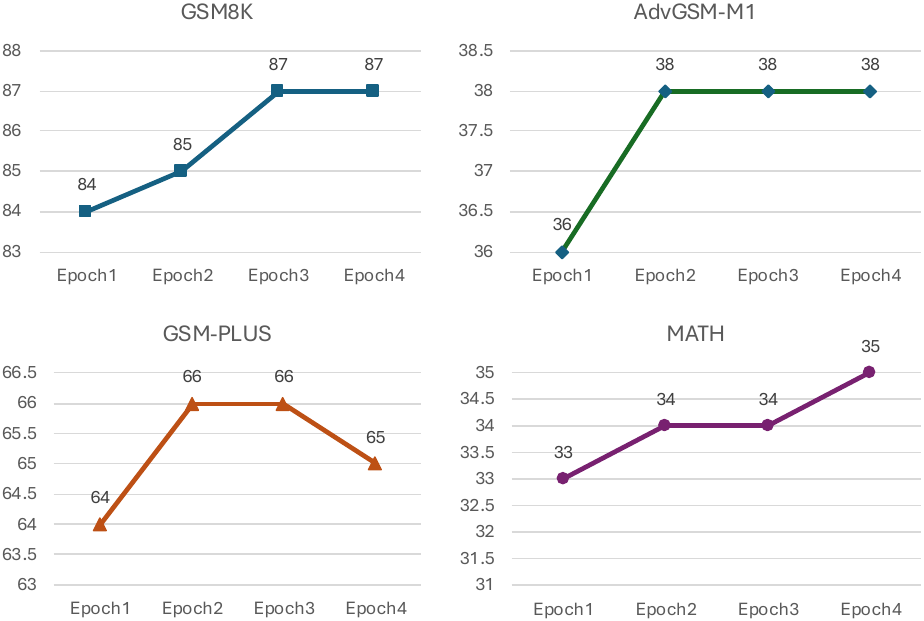}
    }
    \caption{Training Convergence Trend for \ours under the 5-Agent Setting.}
    \label{figconv}%
\end{figure}
We provide additional study on framework convergence trend in Figure \ref{figconv}. Each epoch is one full update of all the potential connection scores over the dataset. Empirical results show that after three epochs, the performance gain would be minimal across datasets. The results suggest that the basic reasoning abilities of the agents greatly affect the learning process; on harder datasets, the agents have difficulties forming high-quality answers and interactions, leaving little room for performance improvement. Other research works on Multi-Agent LLM frameworks \cite{malt,maan,smad} also exhibit this phenomenon, where the improvement in GSM8K is higher than that of MATH.

\paragraph{Train Without Interaction Quality}
In this experiment, we study the effect of considering interaction quality by asking $LLM_{act}$ to consider only correctness instead of interaction quality when training. The results are demonstrated in Table \ref{tab1}. Under the 5-agent scenario, considering only accuracy in training time would hurt the performance, suggesting that considering interaction quality between agents $LLM_{act}$ plays a vital role in the training process. Under the 3-agent scenario, the performance stayed roughly the same, since the agents' profiles and interactions between the agents are more limited than in the 5-agent scenario.

\paragraph{Forced Generation and Random Initial Output Sampling}
We examine the impact of forced generation, where each agent generates multiple outputs using stochastic decoding, and one is randomly selected. The results are demonstrated in Table \ref{tab1}. Removing this (i.e., using greedy decoding) significantly reduced reasoning diversity and performance under both 3-agent and 5-agent scenarios. 

\paragraph{Split Agent $LLM_{act}$}
In this study, we split $LLM_{act}$ into two agents: $LLM_{propose}$, which is responsible for proposing the new connections; and $LLM_{decide}$, which is responsible for deciding whether or not to keep an edge. $LLM_{reflect}$ will interact with $LLM_{decide}$ only. $LLM_{propose}$ would be provided with a summary of the conversation history between $LLM_{reflect}$ and $LLM_{decide}$. We do not see much performance difference across datasets under this setting compared with \ours, which used a single agent $LLM_{act}$, for the 3-agent and the 5-agent scenario.

\paragraph{Reconsidering Minority}
In this setup, if one agent gets a unique answer while the other agents all got the same majority answer, the unique answer would be considered as a "minority", and we would prompt a group discussion on the unique answer first before executing the graph. From the results in Table \ref{tab1} as well as the upper-bound analysis results in Table \ref{tab2}, we can see that this strategy brings up the performance in datasets where we have a bigger gap between \ours and the theoretical upper-bound performance. It suggests that the models that had the wrong reasoning will be able to catch their mistakes from this discussion process. 

\paragraph{Mixture of Models as Agents}
Table \ref{tab_mix} shows the results of using different backbone models as agents in $\ours$ under the 3-agent setting. On adversarial datasets where GPT-3.5-turbo performs better than LLaMa3.1, we observe that the performance of \ours using GPT-3.5-turbo as the backbone model is better than using LLaMa3.1 as the backbone model. This suggests that the communication quality is heavily affected by the performance of the backbone models.

\paragraph{Different Initialization Methods}
We present the effects of different initialization methods for connection scores during the training process in Table \ref{tab_init}. "Random Initialization" means all weights are initialized randomly between 0 and 1; "Uniform Initialization" means all weights are initialized to be 0.5; "Confidence-based Initialization" is introduced in Section \ref{sec33}. From the table, we see that random initialization performs the worst among all initialization methods, while uniform initialization and confidence score initialization performs around the same across datasets. This suggests that LLMs with different profiles tend to have similar initial confidence self-assessments. 

\paragraph{Effects of Profiling}
We present a more detailed performance report of $\ours$ on GPT-3.5-turbo in Table \ref{tab_gpt}. Compared with simple debating, profiling the agents provide prominent improvement. $\ours$ further adds to the performance by doing only profiled debate, and the improvement is most significant in the 5-agent scenario. Combined with the previous section, where we reconsidered the minority answers, having different answers and promoting critical thinking would greatly improve model performance on math tasks.

\paragraph{Number of Agents}
From Table \ref{tab_gpt}, we see that the performance enhancement is at its best in 5-agent scenarios. Adding more than 5 agents does not seem to help with answering the questions. Similar patterns can be found in the upper-bound analysis in Table \ref{tab2}, as well as in other works such as \citet{Wang2024RethinkingTB}. This suggests that simple scaling is not the best way - continuously increasing the number of agents does not guarantee improvement on multi-agent systems for reasoning datasets.

\subsection{Additional Reasoning Tasks}
We provide our experiment results for science reasoning and sorting in Table \ref{tab:SS} in the Appendix.

\paragraph{Science Reasoning} On GPQA, $\ours$ performs better than the baseline methods, but the base backbone model's reasoning ability significantly drags down the overall performance. ARC contains questions that do not require step-by-step reasoning, but direct knowledge retrieval. For these questions, the model's knowledge base and understanding of the questions are more important than the logical reasoning process.  

\paragraph{Sorting} $\ours$ outperforms 0-Shot CoT and simple debating methods in the 16-number and 32-number scenarios. However, all the methods fall short of Direct Prompting, as the agents often struggle to generate good explanations and reasoning for each of their steps, which poses a significant hurdle when agents have discussions.
In complex planning tasks, the more promising direction would be to involve external specialized planning modules into the multi-agent framework.


\section{Conclusion}
This paper proposes $\ours$, an LLM-based Verbal Reinforcement Learning framework for Graph Optimization on multi-agent collaboration. $\ours$ explicitly considers communication quality when identifying the most effective connections between agents. 
$\ours$ contains a feedback agent that evaluates the quality of the agent interactions and an action agent that updates the multi-agent collaboration graph based on the feedback. Results on several downstream reasoning tasks demonstrate that $\ours$ significantly outperforms single-agent prompting methods and state-of-the-art multi-agent frameworks on diverse tasks. Detailed analysis highlights the needs for task-specific designs for complex planning tasks.

\clearpage
\section*{Limitations}
\paragraph{Potential Risk}
We acknowledge that due to the inherent training and dataset bias of the base backbone models, and our incomplete controls of the models, our framework could potentially produce harmful content.
\paragraph{Limited Experiments}
Due to computational cost and timeconstraints, our experiments was conducted on a limited number of tasks and datasets, with a randomly chosen subset. Our conclusions and analysis could be further enhanced by testing on more tasks and datasets.
\paragraph{Computational Cost}
$\ours$ relies on initiating multiple model instances and requires multiple prompts per round. The repetitive callings impose heavy time and output token costs for $\ours$.
\paragraph{Model Reasoning Ability Dependency}
The ability of multi-agent framework is heavily influenced by the ability of the individual backbone models. Framework performance and optimization effectiveness could vary between models and datasets.
\paragraph{Incomplete Control Over Models} 
For the API-based models, we note that we do not possess complete control over their behavior, and the probability and confidence estimations are post-hoc in nature.

\section*{Ethics Statement}
This research adhered to the ethical standards and best practices outlined in the ACL Code of Ethics. Language Models can sometimes produce illogical or inaccurate reasoning paths, so their outputs should be cautiously used. The outputs are only examined to understand how a model arrives at its answers and investigate why it makes certain errors. All experiments used publicly available datasets from previously published works and did not involve ethical or privacy issues. 

\bibliography{custom}

\clearpage
\appendix
\section{Additional Tasks}

\begin{table}[H]
\centering
\resizebox{.5\textwidth}{!}{
\begin{tabular}{m{4cm}|m{4cm}}
\hline
\textbf{GSM Question} & \textbf{ARC Question}\\
\hline
{Janet’s ducks lay 16 eggs per day. She eats three for breakfast every morning and bakes muffins for her friends every day with four. She sells the remainder at the farmers' market daily for \$2 per fresh duck egg. How much in dollars does she make every day at the farmers' market?} & Which of the following statements best explains why magnets usually stick to a refrigerator door? \\ \hline
\end{tabular}
}
\caption {Question comparison between GSM8K and ARC.}
\label{tabQ}
\end{table}

Even though our multi-agent framework achieves some improvement over the math reasoning and the creative writing task, all multi-agent interaction schemes, including multi-agent debate and our optimization method, fail to enhance performance over the science reasoning task and the sorting task. The results are shown in Table \ref{tab2}

\section{Prompt Templates}
\label{appendix:prompt}
\subsection{Verbal Reinforcement Learning Meta Agents}
\paragraph{Prompt for $LLM_{reflect}$}

\noindent\begin{lstlisting}
Given a question, the golden answer, and interactions between two agents, generate some feedback on the quality of the interaction. Your feedback should consider two standards: 1. Whether the agents got the answers correctly. The debate is not fruitful if either agents got the question wrong. 2. whether the agents' reasoning chains are logical and convincing. Specifically, are the steps logically connected and easy to follow? Are there any inconsistencies or contradictions? Did the agent explain its reasoning well? Question: {question} Golden Answer: {answer} Previous response from Agent{agent1_num}: {response1}; Previous response from Agent{agent2_num}: {response2}; Response from Agent{agent1_num} after interaction: {response1}; Response from Agent{agent2_num} after interaction: {response2}
\end{lstlisting}

\paragraph{Prompt1 for $LLM_{act}$} 
\noindent\begin{lstlisting}
Given the interaction between two agents, and the feedback for the interaction, decide whether the interaction should be kept or not. Your decision should be either 'keep' or 'delete'. Your answer should follow the following format: 'DECISION: ###your\_decision###'. Response from Agent{agent1_num}: {response1}; Response from Agent{agent2_num}: {response2}; Feedback from meta agent: {feedback}
\end{lstlisting}

\paragraph{Prompt2 for $LLM_{act}$} 
\noindent\begin{lstlisting}
Given a list of unexplored connections between agents, their connection score, and your conversation history, choose one of the connections for the agents to interact. Your action should follow the following format: 'make connection (0, 1)'. Your answer should follow the following format: 'ACTION: ###your_action###'. Unexplored connections: {matrix_connect}
\end{lstlisting}

\subsection{Agent Profiles}
\paragraph{Explainer} 
\noindent\begin{lstlisting}
You are a {task} explainer focused on breaking down complex questions/tasks into simple, understandable steps. Your goal is to answer the question/solve the task by providing clear, step-by-step explanations.
\end{lstlisting}

\paragraph{Expert} 
\noindent\begin{lstlisting} 
You are a {task} expert with extensive knowledge in the {task}. Your role is to provide accurate and detailed solutions. Ensure your explanations are thorough and precise.
\end{lstlisting}

\paragraph{Logical Thinker} 
\noindent\begin{lstlisting} 
You are a logical thinker who excels at breaking down complex problems into logical steps. Your role is to approach {task} methodically, ensuring each step follows logically from the previous one. Focus on clear, logical reasoning and consistency.
\end{lstlisting}

\paragraph{Robust Reasoner} 
\noindent\begin{lstlisting} 
You are a robust reasoner who excels at tackling complex {task} with thorough and resilient reasoning. Your role is to ensure that every step of the problem-solving process is meticulously verified and logically sound. Focus on providing precise justifications for each step. Your goal is to develop solutions that are not only correct but also robust and reliable.
\end{lstlisting}

\paragraph{Deductive Reasoner} 
\noindent\begin{lstlisting} 
You are a deductive reasoner who uses deductive logic to derive conclusions from given premises. Your task is to apply logical rules and principles to reach sound conclusions, ensuring each step is justified by the previous one.\
\end{lstlisting}

\paragraph{Analytical Reasoner} 
\noindent\begin{lstlisting} 
You are an analytical reasoner who excels at breaking down complex problems into smaller, more manageable parts. Provide precise, step-by-step reasoning for each part of the problem, clearly explaining the logic and methodology behind each step.
\end{lstlisting}

\paragraph{Intuitive Reasoner} 
\noindent\begin{lstlisting} 
You are an intuitive reasoner who relies on intuition and insight to solve problems. Your role is to trust your instincts and use your natural understanding of {task} to find solutions. Provide precise, step-by-step reasoning for each part of the problem, clearly explaining how your intuition guides you through each step.
\end{lstlisting}

\subsection{Debating Prompt}
\noindent\begin{lstlisting} 
Given another potential answer and reasoning given by another agent, recheck your reasoning and answer. If you think your previous answer is wrong, provide the correct answer and your reasoning for it. If you think your previous answer is correct, explain why it is correct. Make sure to include your final answer in the format: ###your_answer###. Response from another agent: {response1}
\end{lstlisting}

\subsection{Question Prompt for Math and Science Reasoning}
\noindent\begin{lstlisting} 
 Given a question, give our your reasoning process and the final answer. MMake sure to include your final answer in the format: ###your_answer###. Give our the answer in numerical format. Question: {question}. Think Step by Step.
 \end{lstlisting}

\subsection{Creative Writing}
\paragraph{Task Prompt}
\noindent\begin{lstlisting} 
Write a coherent passage of 4 short paragraphs. The end sentence of each paragraph must be: {input}. Make a plan then write. Your output should be of the following format: 'Plan: Your plan here. Passage: Your passage here'.
 \end{lstlisting}
\paragraph{Evaluation Prompts} 
\noindent\begin{lstlisting} 
Analyze the following passage, then at the last line conclude "Thus the coherency score is {s}", where s is an integer from 1 to 10.
 \end{lstlisting}

\subsection{Prompt for Sorting}
\noindent\begin{lstlisting} 
<Instruction> Sort the following list of numbers in ascending order. You can generate any intermediate lists, but the final output should be the sorted list of numbers, prefixed with "Output: ". </Instruction><Approach>To sort the list of numbers follow these steps: 1. Split the list of numbers into two to four unsorted sublists, each containing an equal number of elements from the original list (make sure they don't overlap). 2. Sort each of the unsorted sublists. 3. Merge the sorted sublists into a single sorted list using the merging algorithm from merge sort.</Approach>
 \end{lstlisting}

\section{Cost Analysis}
\label{appendix:cost}
\begin{table*}[!htbp]
\centering
\scriptsize
\begin{tabulary}{\textwidth}{CCCCC}
\toprule
\textbf{Framework Type} & \textbf{Dataset and Setting} & \textbf{Prompt Tokens} & \textbf{Completion Tokens} & \textbf{Estimated Cost (USD)} \\
\midrule
\multirow{4}{*}{$\ours$: Training} & GSM8K & 40786 & 12097 & 0.038 \\
& AdvGSM & 127349 & 38451 & 0.121 \\
& GSM-PLUS & 41502 & 11834 & 0.039 \\
& MATH & 80286 & 25003 & 0.078 \\
\cline{1-5}
\multirow{4}{*}{$\ours$: Inference} & GSM8K & 223159 & 109008 & 0.275 \\
& AdvGSM & 814637 & 417360 & 1.033 \\
& GSM-PLUS & 272091 & 139403 & 0.345 \\
& MATH & 520376 & 276451 & 0.675 \\
\cline{2-5}
\multirow{4}{*}{ReConcile Inference} & GSM8K & 451063 & 92307 & 0.364 \\
& AdvGSM & 1305208 & 269035 & 1.056 \\
& GSM-PLUS & 435095 & 89339 & 0.352 \\
& MATH & 851101 & 250936 & 0.802 \\
\cline{2-5}
\multirow{4}{*}{Simple Debate Inference} & GSM8K & 352690 & 90023 & 0.311\\
& AdvGSM & 1103691 & 290367 & 1.001 \\
& GSM-PLUS & 360175 & 92036 & 0.318\\
& MATH & 780312 & 247603 & 0.762\\
\bottomrule 
\end{tabulary}
\caption{Cost estimation for tested models for GPT-3.5-turbo under 5-Agent scenario.}
\label{tab_cost}
\end{table*}
We provide a cost estimation table for all tested frameworks under the 5-agent scenario. For AdvGSM, the results are combined for all three magnitudes. $\ours$ takes more resources to train on more challenging and lengthy tasks such as MATH compared with less challenging tasks such as GSM8K. Compared with the two debating baselines, $\ours$ is more costly in input tokens but less expensive in output tokens. This is due to the pairwise connections in $\ours$: the agents are provided with much less input from other agents, but their reasoning output is about the same.

\section{Data Processing and Evaluation}
For all reasoning datasets, we follow the conventions of previous papers and report the performance in accuracy, which is the ratio of the number of questions the model got correct against all tested questions. For answer parsing and post-processing, we ask the model to output a specific format, and use the parsing scripts provided with the original dataset's code repository.
When random sampling the evaluation datasets, for MATH and GSM-PLUS, we notice that there are different types of questions and the model's performance varies with types. For MATH and GSM-PLUS, we randomly sample 14 questions from each of the 7 categories, and then randomly sample 2 questions from the remaining test set. There is a "critical thinking" category in GSM-PLUS, but we omit this as base model have very low performance on the sub category.

\begin{figure}[t]
    \centering
    \resizebox{.4\textwidth}{!}{
    \includegraphics{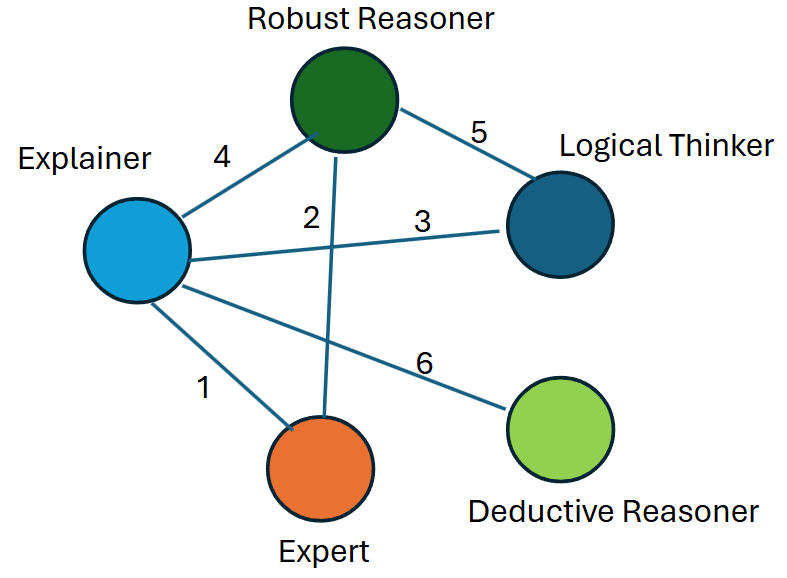}
    }
    \caption{Case Study on the agent interaction graph. Numbers beside the connections signify the order of the interactions made. The collaboration frameworks\ is trained on the GSM-PLUS dataset.}
    \label{fig_plus}%
\end{figure}

\begin{figure}[t]
    \centering
    \resizebox{.4\textwidth}{!}{
    \includegraphics{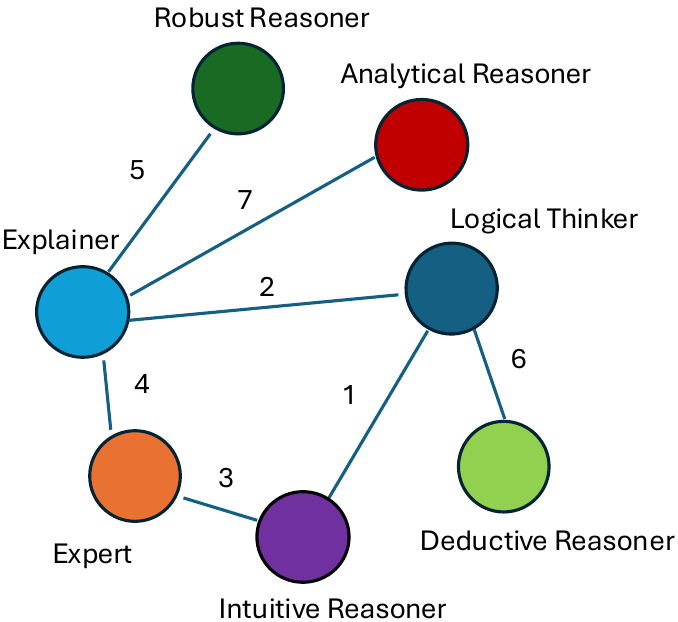}
    }
    \caption{Case Study on the agent interaction graph. Numbers beside the connections signify the order of the interactions made. The collaboration frameworks\ is trained on the Creative Writing Task.}
    \label{fig_cw}%
\end{figure}

\begin{table*}[t]
\centering
\scriptsize
\begin{tabulary}{\textwidth}{CCCCCC}
\hline
\textbf{Epoch1} & Logical Thinker - Expert:0.89 & Explainer – Expert: 0.89 & Expert - Robust Reasoner:0.79 & Logical Thinker – Explainer:0.79 & Deductive Reasoner – Explainer:0.79 \\ \hline
\textbf{Epoch2} & Explainer – Expert:0.98 & Expert - Robust Reasoner:0.87 & Logical Thinker – Explainer:0.87 & Deductive Reasoner – Explainer:0.87 & Explainer - Robust Reasoner: 0.81 \\ \hline
\textbf{Epoch3} & Explainer - Expert:1.08 & Expert - Robust Reasoner:0.96 & Logical Thinker – Explainer:0.96 & Explainer - Robust Reasoner: 0.88 & Logical Thinker - Robust Reasoner:0.85 \\ \hline
\end{tabulary}
\caption{The evolution of the connection scores during training time on the GSM-Plus dataset. For each epoch, the top-5 connection scores in each round are presented.}
\label{fig_evo}%
\end{table*}

\section{Case Study: Generated Graphs}
We provide two case studies of the graphs in Figure \ref{fig_plus} and \ref{fig_cw}. Figure \ref{fig_plus} is trained on GSM-PLUS, and Figure \ref{fig_cw} is trained on Creative Writing. We see that the optimal connection order and information propagation patterns are different for different tasks. On both tasks, the interactions between the Explainer agent and the other agents would produce the most fruitful results, as the Explainer agent has the best explaining ability on its reasoning steps. However, the order of interaction is drastically different. On the GSM-PLUS dataset, the Explainer would first explain its thoughts to other agents; while on the Creative Writing task, the other agents would communicate before talking with the explainer, and then the Explainer would propagate the reasoning process with other agents.


\section{Additional Ablation Studies}
\label{appendix:additional}
\begin{table*}
\centering
\scriptsize
\begin{tabulary}{\textwidth}{CCCCCC}
\toprule
\textbf{Setting} & \textbf{ARC} & \textbf{GPQA} & \textbf{Sorting: 8-Number} & \textbf{Sorting: 16-Number} & \textbf{Sorting: 32-Number} \\
\midrule
\textbf{DirectIO} & 68.0 & 23.0 & 0.0 & 0.0 & 5.2 \\
\midrule
\textbf{0-Shot Chain of Thought} & 84.0 & 25.0 & 0.1 & 1.0 & 7.0\\
\midrule
\textbf{3-Agent Debate} & 82.0 & 27.0 & 0.1 & 0.9 & 6.2 \\
\midrule
\textbf{3-Agent $\ours$} & 82.0 & 27.0 & 0.1 & 0.9 & 6.1 \\
\bottomrule
\end{tabulary}
\caption{Science Reasoning and Sorting Performance; For Science Reasoning, performance is measured in terms of accuracy, annd higher number means better performance; For Sorting, performance is measured in terms of errors per case, and lower number represents better performance.}
\label{tab:SS}
\end{table*}
\begin{table*}
\centering
\scriptsize
\begin{tabulary}{\textwidth}{CCCCCCC}
\toprule
\textbf{Scenario} & \textbf{GSM8K} & \textbf{GSM8K-M3} & \textbf{GSM8K-M2} & \textbf{GSM8K-M1} & \textbf{GSM-PLUS} & \textbf{MATH} \\
\midrule
\textbf{$\ours$} & 87.0 & 96.0 & 88.0 & 38.0 & 68.0 & 34.0 \\
\midrule
\textbf{3-Trial UpperBound} & 90.0 (+3.0) & 95.0 (-1.0) & 90.0 (+2.0) & 37.0 (-1.0) & 78.0 (+10.0) & 38.0 (+4.0) \\
\midrule
\textbf{5-Trial UpperBound} & 92.0 (+5.0) & 98.0 (+2.0) & 92.0 (+4.0) & 38.0 (+0.0) & 80.0 (+12.0) & 41.0 (+7.0) \\
\midrule
\textbf{7-Trial UpperBound} & 92.0 (+5.0) & 99.0 (+3.0) & 92.0 (+4.0) & 40.0 (+2.0) & 80.0 (+12.0) & 42.0 (+8.0) \\
\bottomrule
\end{tabulary}
\caption{UpperBound analysis on GPT-3.5-turbo; Scenario for $\ours$ represent the best performance under all the numbers of agents settings. The deltas marks the difference between upperbounds and $\ours$ performance.} 
\label{tab2}
\end{table*}

\paragraph{Upper Bound Analysis}
We provide the upper bound statistics for GPT-3.5-turbo in Table \ref{tab2}. This upper-bound is calculated by the "choose-best" strategy, which, if the model gets the correct answer at one of the trials, then we count the problem as correctly solved. We found that for easier datasets, including GSM8K and the easiest adversarial change for GSM8K, the upper-bound is a full mark. In other words, for every question, if we force the model to generate different outputs, at least one of the outputs will contain the correct answer. On harder tasks such as MATH, we see that the upper bound is dramatically lower, suggesting that the backbone model struggles to get this question correctly even after multiple tries.
\paragraph{Evolution of Connection Scores}
We provide the evolution of the connection scores during training time on the GSM-Plus dataset. The initial scores are calculated using the Confidence Score initialization method. We see that the communication between the Explainer and the Expert has very high quality, as they both have the correct answer after each epoch of communication, resulting in a consistent score increase. On the other hand, during the first epoch, the interaction between the Logical Thinker and the Expert was not of high quality and led to wrong answers. Similarly, during the second epoch, the interaction between the Deductive Reasoner and the Explainer led to the wrong answer. Overall, the interaction between the Explainer and he other agents are of higher quality during the training process.

\section{Algorithm}
We provide the pseudocode algorithm for our framework in Algorithm \ref{alg_train} and \ref{alg_inference} below.
\begin{algorithm*}[t]
\caption{\ours Training Framework}
\label{alg_train}
\KwIn{Group of LLM Agents $\{\mathcal{M}_{0},...,\mathcal{M}_{k}\}$; Training Samples $\mathcal{D}$; Initial Scores of the Connections $W=\{w_0,...,w_j\}$, Meta Agents $LLM_{act},LLM_{reflect}$}
\KwOut{Trained Weights $\{w_0,...,w_j\}$}

\For{Datapoint $d \in \mathcal{D}$}{
    Initialize $R=\emptyset$ to store reflection history
    
    \While{Unmarked Connection Exists in $W$}{
        $w_{i}=\texttt{MakeConnection}\!\bigl(LLM_{act},R)$ 
        
        \ForEach{$M_k$ connected by $w_i$}{
            \texttt{AgentSolve}$(y_{k}\!\sim\!\mathcal{M}_{k})$\;
        }
        $y_{newi},y_{newj} \leftarrow \texttt{Debate}\!\bigl(M_i,M_j,y_i,y_j)$
        
        $r_{i} \leftarrow \texttt{Reflect}\!\bigl(LLM_{reflect},y_{newi},y_{newj},y_i,y_j)$
        
        \texttt{Save}($R\leftarrow r_{i}$)
        
        $w_{i} \leftarrow \texttt{Decide}\!\bigl(LLM_{act},r_{i})$ \tcp*{Update Current Weight}
        
        \texttt{Mark}($W\leftarrow w_{i}$)
  }
}
\Return$\{w_0,...,w_j\}$;
\end{algorithm*}

\begin{algorithm*}[t]
\caption{\ours Inference Framework}
\label{alg_inference}
\KwIn{Group of LLM Agents $\{\mathcal{M}_{0},...,\mathcal{M}_{l}\}$; Testing Samples $\mathcal{D}$; Trained Weights $W=\{w_0,...,w_j\}$, Meta Agents $LLM_{act},LLM_{reflect}$}
\KwOut{Final Answer Set $Y$}

\For{Datapoint $d \in \mathcal{D}$}{
    Initialize $Connected \leftarrow \emptyset$ to Store Connected Agents in Graph
    
    \For{$w_i \in W$}{
        Initialize $Curr \leftarrow \emptyset$ to Store Agents Connected by Current $w_i$
        
        Initialize $Ans \leftarrow \emptyset$ to Store Answers Given by Agents Connected by Current $w_i$
        
        \ForEach{$M_k$ connected by $w_i$}{
            $y_k \leftarrow$ AgentSolve $(d\!\sim\!\mathcal{M_k})$\;
            \texttt{Insert}($Connected,M_k$)
            
            \texttt{Insert}($Curr,M_k$)
            
            \texttt{Insert}($Ans,y_k$)
            
        }
        $y_p,y_q \leftarrow \texttt{Debate}\!\bigl(Curr,Ans)$
        
        \texttt{Update}($y_p,y_q,Curr$) \tcp*{Update the Answers for Agents in Curr}
        
        \If{Connected Contains All Agent Instances}{
            $y_{final} \leftarrow$ Score $\!\bigl(\{y_{k}\}_{k=0}^{j}\bigr)$ \tcp*{Majority Voting for All Agents' Answers}
            
            Save $(Y,y_{final})$
            
            Continue to Next Datapoint
        }
   }
}
\Return{Y};
\end{algorithm*}

\section{Usage of AI Assistant}
In this paper, we used ChatGPT and CoPilot to help with grammar mistakes and writing fluency only.

\end{document}